# Using Machine Learning-Based Models for Personality Recognition

**Fatemeh Mohades Deilami[1], Hossein Sadr[2,*] , Mozhdeh Nazari[3]**

[1] Department of Computer Engineering, Ayandegan Institute of Higher Education, Tonekabon, Iran; Deilami@aihe.ac.ir.
[2] Department of Computer Engineering, Rahboard Shomal Institute of Higher Education, Rasht, Iran; Sadr@qiau.ac.ir.
[3] Cardiovascular Diseases Research Center, Guilan University of Medical Sciences, Rasht, Iran; Mojdeh.nazari@qiau.ac.ir.

## Abstract

Personality can be defined as the combination of behavior, emotion, motivation, and thoughts that aim at describing various aspects of human behavior based on a few stable and measurable characteristics. Considering the fact that our personality has a remarkable influence in our daily life, automatic recognition of a person's personality attributes can provide many essential practical applications in various aspects of cognitive science. Although various methods have been recently proposed for the task of personality recognition, most of them have mainly focused on human-designed statistical features and they did not make use of rich semantic information existing in users' generated texts while not only these contents can demonstrate its writer's internal thought and emotion but also can be assumed as the most direct way for people to state their feeling and opinion in an understandable form. In order to make use of this valuable semantic information as well as overcoming the complexity and handcraft feature requirement of previous methods, a deep learning based method for the task of personality recognition from text is proposed in this paper. Among various deep neural networks, Convolutional Neural Networks (CNN) have demonstrated profound efficiency in natural language processing and especially personality detection. Owing to the fact that various filter sizes in CNN may influence its performance, we decided to combine CNN with AdaBoost, a classical ensemble algorithm, to consider the possibility of using the contribution of various filter lengths and gasp their potential in the final classification via combining various classifiers with respective filter size using AdaBoost. Our proposed method was validated on the Essay dataset by conducting a series of experiments and the empirical results demonstrated the superiority of our proposed method compared to both machine learning and deep learning methods for the task of personality recognition.

## 1 | Introduction

Personality is a stable tendency and characteristic that determines the similarities and differences in psychological behaviors (thoughts, emotions, and actions) of individuals. In other words, personality not only shows an individual's behavioral pattern, thought, and interpersonal communications but also has a great impact on various life aspects, such as happiness, preference, physical, and mental health [1] and [2]. The evolution of personality theories and their progress in measurement methods and statistical analysis has led to the emergence of one of the most influential contemporary personality theories, known as the Five Factor Model (FFM) or Big Five, which contains five primary traits and has been considered by many psychologists in recent years as a popular and powerful approach for studying personality traits [3] and [4]. According to the FFM, personality consists of five main dimensions including neuroticism (NEU), extraversion (EXT), openness to experience (OPN), agreeableness (AGR), and conscientiousness (CON). It is worth mentioning that the study of personality is not only essential for psychology and personality recognition but also can benefit various applications, such as cognitive science [5], social network analysis [6], sentiment analysis [7] and [8], recommender systems [9], deception detection [10], and so on.

Considering the importance of automatic personality recognition, numerous studies have been conducted in this field over the past few years [11]. In other words, although texts generated by individuals can be considered as the most direct way of expressing their thought and emotion while they contain rich self-disclosed personal information that is highly correlated with people's personality and interpreting them can provide us with valuable information about users' behavior and feelings, most of the existing methods commonly utilized questionnaire investigations or expert reviews that were not only costly, time-consuming, and less practical but also were highly dependent on an expert and only utilized human-designed statistical features to perform recognition and did not consider the valuable information existing in texts [12].

By taking the significance of textual data into account, a small number of studies have focused on using text generated by people to predict their personality [13]. In this regard, machine learning based methods have been also utilized but their obtained results were not satisfactory because the majority of them were based on statistical or hand-craft linguistic features and were not able to consider the rich user-generated textual information and extract features from them automatically while these words and text are the most valuable features for determining the emotion and personality [14].

By the development of deep neural networks, they demonstrated remarkable performance in various Natural Language Processing (NLP) tasks including opinion mining and sentiment analysis [7] and [8]. It must be noted that personality recognition is very similar to NLP applications while they both focus on mining users' attributes from texts. Accordingly, employing powerful text modeling techniques that have been efficiently utilized in the NLP domain can be the most intuitive and straightforward idea for improving the performance of personality recognition [12].

Having the mentioned limitations besides the potential of deep learning in our mind, we proposed a deep learning based method for personality recognition that tries to make use of both Convolutional Neural Network (CNN) and AdaBoost algorithm [15]. Although CNN has been successfully utilized for various NLP tasks and extracting local features can be considered as its potential, using various filter lengths may have a negative influence on the efficiency of the CNN classifier. To this end, we decided to combine CNN with AdaBoost algorithm to investigate the possibility of leveraging the contribution of different filter lengths and gasp their potential for personality recognition by combining classifiers with respective filter sizes. The reason behind choosing AdaBoost is that it is a Meta algorithm that can be used in conjunction with other learning algorithms to improve classification accuracy. Based on this algorithm, the classification of each new stage is adjusted in favor of incorrectly classified samples in the previous stages. In fact, with the help of AdaBoost algorithm, the classification process is repeated until the classification error is minimized. The proposed method was conducted on James Pennebaker and Laura King's stream-of-consciousness Essay dataset [16]. In summary, the contributions of this paper are as follows:

− *We designed a new structure based on the integration of CNN and AdaBoost for predicting personality from texts where different lengths and various weigh matrices are used in the convolutional layer to extract features.*
− *We considered different variations of the proposed method in our implementation to prove the efficiency of vector representation in personality recognition.*
− *To the best of our knowledge, it is the first study that tries to investigate the combination of CNN and AdaBoost besides the importance of vector representation for personality recognition. Based on the empirical result, the proposed method demonstrated higher efficiency compared to both machine learning and deep learning based methods.*

The remainder of this paper is organized as follows: Related studies with a focus on deep learning based methods are presented in Section 2. Section 3 includes the details of the proposed methods. Experimental details and obtained results are extensively reported in Section 4. Conclusion and possible future directions are also mentioned in Section 5.

## 2 | Related Work

Personality theorists have developed unique methods for assessing individuals' personalities. By applying these methods, they obtained valuable information and then based their formulations on it. Personality recognition methods can be generally divided into two major categories of psychology based and artificial intelligent based methods [12]. Given that the focus of this paper is on automatic personality recognition from texts, especially people's opinion about various topics, the present study falls into the group of artificial intelligent based methods. Artificial intelligent based methods for personality recognition are divided into two groups of machine learning and deep learning based methods. More details about these methods and related studies are reported in the following.

### 2.1 | Machine Learning Based Methods

By the rapid development of the Internet and social media, numerous studies have been conducted on personality recognition from text. In this regard, Golbeck et al. [17] used M5' rule and Gaussian process for predicting personality based on Big Five scores. They utilized 167 Facebook users' personal information, activity, preferences, and language usage to extract 77 features and perform classification. Following a similar line of research, Golbeck et al. [18] utilized 297 Twitter users' information and applied a similar approach to predict the personality. Moreover, the relation between personality and various kinds of users was also analyzed by Quercia et al. [19]. They used M5' algorithm to predict 335 Twitter users' personalities based on Big Five attributes according to their number of followers, following, and listed counts. Alam et al. [20] utilized a bag of word methods besides unigrams as features to perform personality recognition. They applied various techniques, such as Support Vector Machine (SVM), Multinomial Naïve Bayes (MNB), and Bayesian Logistic Regression (BLR) to predict Big Five attributes according to my personality corpus. Skowron et al. [21] collected text, image, and users' metadata from Instagram and Facebook and performed various machine learning techniques to predict personality. They concluded that joint analysis could enhance performance. Li et al. [22] also proposed a semi-supervised method that utilized over 547 Chinese active users of SinaWeibo to predict personality.

In the following, Bai et al. [23] utilized the information of 209 users of PenPen (Chinese social network) to predict their personality. They analyzed various attributes including usage statics, emotional state, and demographic information, and then applied C4.5 decision tree to perform classification. Peng et al. [24] utilized SVM to predict the personality of 222 Chinese Facebook users based on their generated texts about various topics. Argamon et al. [25] used word categories and relative frequency of function words as SVM input to make discrimination between students at the opposite extremes of neuroticism and extraversion. The efficiency of various textual features extracted from the psycholinguistic dictionary or psychologically oriented text analysis tools was also explored by Mairesse et al. [26] N-gram frequency was another feature that was commonly used as the input of SVM or Naïve Bayes for classifying low and high scoring blogger for Big Five personality attributes.

As it is clear, a large number of studies have focused on traditional machine learning methods to perform personality recognition and the majority of them were highly dependent on the handcraft features like online activities, profile information, or manually extracted features of texts. In other words, the machine learning based methods required an expert to extract features and were not able to make use of rich features existing in the text.

### 2.2 | Deep Learning Based Methods

By the rapid growth of deep learning, deep neural networks obtained remarkable efficiency in various NLP tasks. Due to the fact that personality recognition from text is very similar to other NLP tasks like text classification or sentiment analysis, deep neural networks have also found their way in personality recognition. In this regard, Yu and Markov [27] utilized deep learning methods to predict the personality of Facebook users. They applied fully connected network, CNN and Recurrent Neural Network (RNN) in their experiments and proved the superiority of deep learning based methods for personality recognition compared to other existing methods. In similar research, Tandera et al. [28] utilized multilayer perceptron, Long Short Term Memory (LSTM), Gated Recurrent Unit (GRU), and 1-DCNN to predict Facebook users' personality according to Big Five personality attributes. Similarly, Majumder et al. [4] employed CNN for extracting deep semantic features and predicting the personality based on them. Xue et al. [3] also proposed AttCNN model to extract deep semantic features from users' post text and then concatenated them with statistical linguistic features. They fed the obtained features to a regression algorithm to predict the personality based on the Big Five personality attributes. It is worth mentioning that although deep neural networks have been rarely employed for the task of personality recognition, they have obtained considerable results and they are actually in the early steps of their development and growth.

## 3 | Proposed Methodology

Due to the fact that language is the most reliable way for people to state their opinion and internal feeling in an understandable way, it can be considered as a valuable knowledge for psychologists to interpret people's feelings and predict their personality. In other word, while text can reflect various aspects of its author, efficiently modeling the text generated by authors can improve the performance of personality recognition [29]. Motivated by this intuition, we decided to use a reinforced CNN architecture with various filters to perform classification. Unlike previous CNN based methods recognition that combined the properties of different filters into a unified vector that was then fed to a fully connected network to predict the personality [3] and [4], various features obtained from various filters of the CNN are fed to a separate pooling and classifier in our proposed method. When the initial results are obtained by each of the classifiers, the AdaBoost algorithm [15] is used to produce the overall classification results. The reason for choosing AdaBoost is that it is a Meta algorithm that can be used in conjunction with other learning algorithms to improve performance. In this algorithm, the classification of each new sample is adjusted in favor of incorrectly classified samples of other classifiers. In other words, AdaBoost is able to combine weak classifiers with a strong classifier because it can learn the classification error of each weak classifier and accordingly adjust the weight of the classifier for the final classification.

The proposed method includes five steps and it is a combination of CNN and Adaboost algorithm [15]. In the proposed method, different filters with various sizes are used to scan the input sentence and extract precious low-level features from the input text. In other words, each CNN has its own unique convolutional, pooling, and classification layer, and classification is performed separately in each CNN. Finally, the AdaBoost aggregation algorithm is used to create a robust classification based on the different weights of the different classifiers and the personality type is then estimated based on it.

In general, the idea of this paper is based on this hypothesis that using various filters with CNN leads to the generation of different features that each of them may have a different effect on the final classification. In the classical CNN, these features are merged after applying the pooling operation and classification is performed on the merged features. This can eliminate the impact of some features that may contain valuable information. In this regard, we decided to feed the features obtained from each filter to a separate pooling and classification layer. Next, the classification results are then combined using the AdaBoost algorithm to obtain the best final.

### 3.1 | Representation Layer (Word Matrix Formation)

Data representation refers to language modeling techniques in NLP that aims to map words from a very large space to a continuous vector space with much smaller dimensions. In other words, in order to be able to apply a deep learning method for the task of text classification, the words must be transformed into high dimensional vectors to capture the syntactic, semantic, and morphological information of the words. To this end, Skip-Gram [30] model which is a shallow two layers neural network and tries to learn vector representation of a word based on its context is used in the first layer of our proposed method. The goal of the Skip-Gram model is to find word representations that are efficient for the prediction of the surrounded words in a sentence. Let $x_1, x_2 \cdots x_n$ is a sequence of training words, Skip-Gram aims to maximize the average log probability (*Eq. (1)*).

$$\frac{1}{n}\sum_{i=1}^{n} \sum_{-c \leq j \leq c, j \neq 0} \log p(x_{i+j}|x_i). \tag{1}$$

Where $c$ refers to the training context size while a larger value of $c$ yields to more training samples and higher accuracy. Noteworthy, the value of $p(x_{i+j}|x_i)$ is also obtained using SoftMax function Finally, considering that $V$ is the size of the vocabulary and $d$ is the size of word embedding, each word is encoded by a column vector in $A \in \mathbb{R}^{d \times V}$ as the sentence matrix.

## 3.2 | Convolutional Layer

The objective of the convolutional layer is to extract local features as well as retaining the sequential information of the input text. To this end, the obtained sentence matrix $A \in \mathbb{R}^{d \times n}$ is fed to the convolutional layer to produce new features. According to the fact that the sequential structure of a sentence has an important effect in specifying its meaning, it is sensible to choose filter width equal to the dimensionality of word vectors ($d$). In this regard, only the height of filters($h$), known as region size, can be varied.

Considering $A \in \mathbb{R}^{d \times n}$ as a sentence matrix, convolution filter $w \in \mathbb{R}^{h \times d}$ is applied on $A$ to produce its submatrix as a new feature $A[i:i]$. As the convolution operation is applied repeatedly on the matrix of $A$, $O \in \mathbb{R}^{n-h+1 \times d}$ as the output sequence is achieved (*Eq. (2)*).

$$O_i = w \cdot A[i:i+h-1]. \tag{2}$$

Here $i = 1, \ldots, n - h + 1$ and · is the dot product between two matrices of the convolution filter and input submatrix. Bias term $b \in \mathbb{R}$ and an activation function are also added to each $O_i$. Finally, feature maps $C \in \mathbb{R}^{n-h+1}$ are generated (*Eq. (3)*).

$$C_i = f(O_i + b). \tag{3}$$

## 3.3 | Pooling Layer

While various feature maps according to different filter sizes are generated, a pooling function is required to induce fixed size vectors. Various strategies such as average pooling, minimum pooling, and maximum pooling can be used for this aim and the idea behind them is to capture the most important feature from each feature map and reduce dimensionality. Maximum pooling is used in our proposed method (*Eq. (4)*).

$$c_{max} = \max\{C\} = \max\{c_1, \ldots, c_{n-h+1}\}. \tag{4}$$

It is worth mentioning that the pooling layer makes the proposed method aware of the order of the sentences and distributes information related to the personality of the individuals throughout the sentence. On the other hand, the pooling layer allows us to work with sentences of variable length, given that the number of features in the proposed method is aligned with the number of filters. Moreover, the pooling layer reduces the size of the feature maps and future computations. Features obtained from the pooling layer are then processed using a nonlinear function before being classified.

## 3.4 | Regularization Layer and Softmax

In order to overcome overfitting, which is known as one of the most important weaknesses of neural networks, dropout is used as a regularization technique in our proposed method. Based on this technique, the values of some features are set to zero. It means that if $O_{out} = \{O_{out}^1, O_{out}^2, \ldots O_{out}^P\}$ are features obtained from the previous layer ($P$ is the number of filters in the convolutional layer), some of them are randomly set to zero before the SoftMax layer. Notably, the dropout value is a hyper-parameter that is specified along with the training. The classification result is the SoftMax output after the regularization layer. This layer employs the regularized features as the input of SoftMax layer to calculate the probability of distribution over all five different kinds of personality based on Big Five attributes (*Eq. (5)*).

$$P(y=j|x) = \text{softmax}_j(x^T w + b) = \frac{e^{x^T w_j + b_j}}{\sum_{k=1}^{K} e^{x^T w_k + b_k}}. \tag{5}$$

Where $w$ is the input weigh, $b$ is the bias term, $y$ refers to the output class and $K$ is the number of output classes.

## 3.5 | Adaboost Training Layer and Prediction Integration

AdaBoost is an algorithm that tries to integrate weak classifiers into a strong classifier. Accordingly, we employed this algorithm in our proposed method to find the appropriate weights for the classifiers adjusted to different N-grams. In this regard, there is a need to obtain the statistics of the weak classifier results based on the training samples and then adjust the weights of training samples and classifiers to achieve the final strong classification. Backpropagation is also used to train the network ahead of the AdaBoost integrating part. Training process of AdaBoost can be stated as follows:

1. Initialize equal distribution of $D^1$ to all training samples while $D^t$ specifies the ith training sample distribution (*Eq. (6)*).

$$D_i^1 = \frac{1}{\#training\_samples}. \tag{6}$$

2. In each training epoch of $t$:

While backpropagation is applied to train three neural networks consecutively, the following process is performed on all classifiers in each epoch.

## 3.6 | Estimating Weak Classifiers Statistics

After training the classifiers and predicting the output labels, classification statistics over the samples are saved, and weak classifier error $e_m(t)$ in then calculated (*Eq. (7)*).

$$e_m^t = \sum_i D_i^t 1(G_m(x) \neq y(x)). \tag{7}$$

## 3.7 | Adjusting Weight

As a weak classifier is trained, the classification error is used to modify the distribution over the training set. Thereafter, the error index and weak classifiers' weights are calculated.

Calculating classifier weights (*Eq. (8)*):

$$a(m) = \frac{1}{2} \ln \frac{1 - e_m^t}{e_m^t}. \tag{8}$$

Adjusting distribution (*Eq. (9)*):

$$\omega_i^{t+1} = \frac{\omega_i^t \exp(-a(m)y(x)G_m(x))}{\omega^t}. \tag{9}$$

## 3.8 | Improved Validation

As the training process is finished, element-wise multiplication of weights and outputs is performed to obtain the final predicted class of the personality. The learned weight $\omega$ is then used to perform the improved validation using the following equation (*Eq. (10)*) where $i$ is the classifier index, $l$ refers to the output label of the classifier and $a$ determines the ensembles of the classifier weights.

$$L(s) = \sum_i a(i) * l(i). \tag{10}$$

# 4 | Experiments

## 4.1 Dataset

To evaluate the efficiency of the proposed method, various experiments were carried out on the Essay dataset [16] as a standard benchmark. Essay is a large dataset based on the stream of consciousness that was collected by Pennebaker and Laura King according to the text generated by 2467 users between 1997 and 2004 that were labeled based on classes of personality traits including Neuroticism (NEU), Extraversion (EXT), Openness to experience (OPN), Agreeableness (AGR), and conscientiousness (CON)). Therefore, the dataset includes a label for each essay indicating the personality of its author and can be suitable for supervised learning. It is worth mentioning that the texts of this dataset were generated by students of the American Psychological Association. More details about the used dataset and the number of samples in each class are provided in *Table 1*.

Table 1. Number of samples in the essay dataset based on the personality labels.

| Personality Label | Number of positive samples | Number of negative samples |
|---|---|---|
| Neuroticism (NEU) | 1234 | 1234 |
| Extraversion (EXT) | 1191 | 1191 |
| Openness to Experience (OPN) | 1196 | 1196 |
| Agreeableness (AGR) | 1157 | 1157 |
| Conscientiousness (CON) | 1214 | 1214 |

## 4.2 | Evaluation Metrics

Evaluation metrics explain the performance of a method while an important aspect of evaluation metrics is their capability to discriminate among method results. Evaluation of a deep learning based method is generally conducted by comparing the actual labels of training samples with those that are empirically labeled and choosing the best evaluation metric is highly dependent on the task. Noteworthy, the standard metric of Accuracy (*Eq. (11)*) is used in our experiments to perform the evaluation. Where TP, TN, FP, and FN respectively refer to the true positive, true negative, false positive, and false negative.

$$\text{Accuracy} = \frac{TP+TN}{TP+FP+FN+TN}. \tag{11}$$

## 4.3 | Experiment Description

To provide a comprehensive understanding of the efficiency of the proposed method, various experiments with several variations of the proposed method were conducted that are introduced in the following:

– *CNN-AdaBoost-Rand= Random initialized word vectors are used as the input of the proposed method.*
– *CNN-AdaBoost-Static= Pre-trained word vectors obtained from Skip-Gram model are used as the input and their weights are not updated along the training process.*
– *CNN-AdaBoost-Non-Static= Pre-trained word vectors obtained from Skip-Gram model are used as the input and their weights are updated along the training process.*
– *CNN-AdaBoost-2channel= Combination of random initialized word vectors and pre-trained word vectors obtained from Skip-Gram model is used as the input of the proposed method.*

## 4.4 | Model Configuration and Hyper-Parameters

Due to the fact that deep neural networks require a large number of training samples to be accurately trained, a typical processor cannot be expected to perform this operation. Therefore, it is necessary to provide a powerful processor with high speed to perform the training process. All implementations of this paper were conducted on the system with Intel Xeon 2 E5-2620 2.0 GHz processor and 8 GB of RAM using Python as the programming language in the Linux environment.

The implementation process started with preprocessing the input data. To this end, the text was split into the sequence of sentences, and the period and question mark characters. Thereafter, the sentences were split into words. All letters were then reduced to lowercase all characters other than ASCII letters, exclamation marks, digits, and quotation marks were removed. Due to the fact that some essays in the dataset did not include period which yielded to absurdly long sentences, sentences longer than 150 words were split into sentences with 20 words (expect the last piece that could be shorter).

In the following, in order to be able to use words as the input of the proposed method, they must be converted to vectors. In this regard, we trained Skip-Gram model using all existing documents while window size and word vector dimensions were respectively about 5 and 150. Notably, the learning rate of 0.025 was employed to update word vectors and minimize the loss function.

After updating the word vectors, they were fed to the convolutional layer where various filer size (3, 4, and 5) was selected and the number of filters was about 150. Rectified linear function (ReLU) was also utilized as an activation function to apply nonlinearity.

The proposed neural network was trained using the training data and the output classes were determined for the multiplicity of classes by Softmax function. In order to evaluate the output of the experimental data, a cost function was adjusted and ADADELTA update rule was employed for stochastic gradient descent with a learning rate of 0.01 while mini-batch size and dropout rate were respectively about was 25 and 0.05. 60 epochs were also used for training. Hyper-parameters' values used for training of the proposed method are briefly reported in *Table 2*.

Table 2. Configuration of the proposed method hyper-parameters.

| Hyper-parameters | Value |
|---|---|
| Window size | 50 |
| Word vector dimension | 150 |
| Skip-Gram learning rate | 0.025 |
| Filter size | 3,4,5 |
| Number of filters | 150 |
| Activation function | ReLU |
| CNN learning rate | 0.01 |
| Dropout rate | 0.05 |
| Mini-batch size | 25 |
| Number of epochs | 60 |
| Update rule | ADADELTA |

It is worth mentioning that due to the fact that the goal of this paper is to predict the personality based on Big Five traits which include five different classes, we built five different neural networks with the same introduced structure. Moreover, all training and testing experiments were carried out based on 5-fold cross-validation and the whole dataset was randomly divided into five chunks while three chunks were used as the training set and the other two chunks were used as validation and test sets. Notably, the average accuracy of each variation of the proposed method over the 5-fold cross-validation is reported in the result section.

## 4.5 | Results and Analysis

The average testing results over 5-fold cross-validation obtained by different variations of the proposed method in comparison to other existing methods for the task of personality recognition on the Essay dataset are reported in *Table 3*. As it is clear, al variations of the proposed method have superior performance compared to both machine learning and deep learning methods which can be due to the employment of AdaBoost algorithm that aims to separate the features by parsing the documents form different filters and boosting the classifier performance on these representations and modulate them to obtain higher performance. Notably, although AdaBoost algorithm is sensitive to noisy and outdated data, it is superior to most of the learning algorithms in terms of overfitting problems.

Among all variations of the proposed method, it is obvious that CNN-AdaBoost-Rand, which used random initialized word vectors as input, has the lowest accuracy while other variations that employed pre-trained word vectors as input performed slightly better which can be owing to the utilization of Skip-Gram model for proving the rich vector representation.

Table 3. Accuracy comparison of automatic classification of texts in essay dataset based on the big five dimensions of personality.

| | method | EXT | NEU | AGR | OPN | CON |
|---|---|---|---|---|---|---|
| Published state of the art | TF-IDF [31] | 38.13 | 32.96 | 40.35 | 36.24 | 39.11 |
| | N-Gram [31] | 51.74 | 50.32 | 53.14 | 50.17 | 51.07 |
| | MairesseBaseline (MB) [4] | 55.46 | 58.23 | 54.93 | 54.63 | 60.48 |
| | MairesseBaseline+CoarseAff [32] | 56.45 | 58.33 | 56.03 | 56.73 | 60.68 |
| | CNN[4] | 55.73 | 55.80 | 55.36 | 55.69 | 61.73 |
| | RNN [33] | 56.68 | 56.41 | 55.97 | 56.04 | 62.75 |
| | CNN+ Mairesse [4] | 58.09 | 57.33 | 56.71 | 56.71 | 61.13 |
| | RNN+Mairesse [33] | 59.73 | 60.23 | 57.81 | 58.27 | 63.44 |
| Proposed method | CNN-AdaBoost-Rand | 60.05 | 60.91 | 58.11 | 59.34 | 63.94 |
| | CNN-AdaBoost-Static | 60.12 | 61.05 | 58.35 | 59.68 | 64.05 |
| | CNN-AdaBoost-Non-Static | 60.45 | 61.71 | 58.61 | 60.01 | 64.18 |
| | CNN-AdaBoost-2channel | 61.25 | 61.93 | 59.02 | 60.16 | 64.63 |

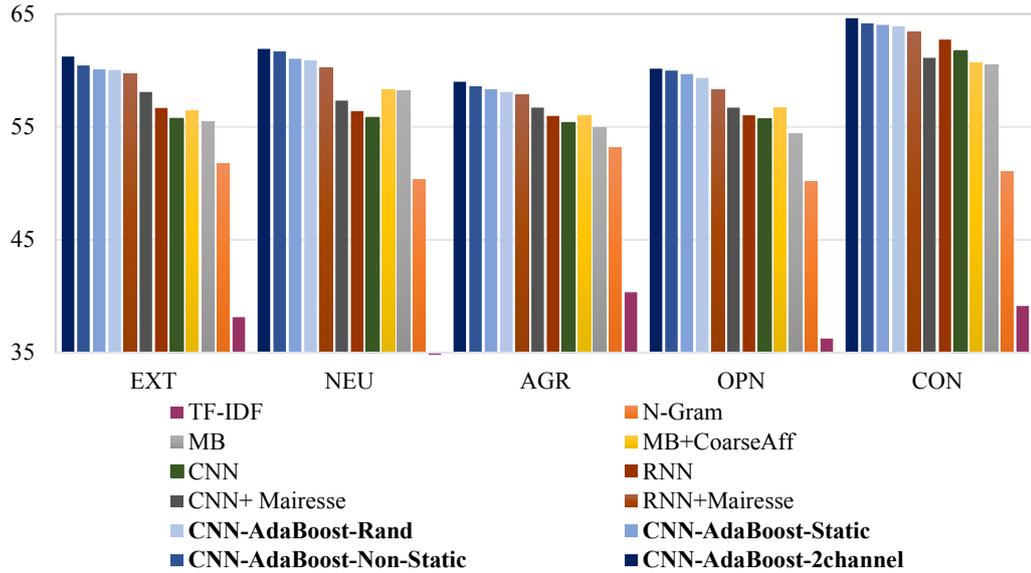

Fig. 5. Diagram of accuracy comparison of automatic classification of texts in essay dataset based on the big five dimensions of personality (all variations of the proposed method are illustrated with blue color).

Moreover, besides CNN-AdaBoost-Rand, CNN-AdaBoost-Static has the lowest accuracy which indicates that updating weight along the training process can enhance the performance. Overall, CNN-AdaBoost-2channel proved to have the highest performance for the task of personality recognition while its obtained accuracy was respectively about 61.25%, 61.93%, 59.02%, 60.16% 64.63 % for extraversion, neuroticism, agreeableness, openness to experience, and conscientiousness. The chart indicating the obtained results is also illustrated in *Fig. 5*.

## 5 | Conclusion

Personality recognition is known as one of the most interesting and practical topics in psychology. Recently, due to the penetration of the Internet in society and the combination of human relations and artificial intelligence, personality recognition has attracted considerable attention in cognitive science. Commonly, machine learning based methods have been successfully utilized for this aim. However, they are confronted with some limitations and are highly dependent on human experts for extracting appropriate features. On the other hand, by the development of deep learning, as a special type of machine learning methods, they presented significant results in NLP natural language processing, particularly personality recognition, due to their amazing capabilities in automatically extracting features and their unique structure. In this regard, CNN can extract low-level features from the text and have been efficiently used for text classification. It is worth mentioning that despite the considerable performance of CNN, they are still facing with some major problems. The prominent challenge of these networks is that is the features, generally n-grams, obtained from various filter sizes can play a different role in the final decision. This means that a 5-gram may extract more relevant information than a 4-gram about the meaning of a sentence. To fill this lacuna, we decided to give the features obtained from various filters of the CNN to separate pooling and classification layers. Therefore, when the initial results were obtained by each of the classifiers, the AdaBoost algorithm was used to produce the overall classification results. The goal of AdaBoost algorithm is to increase the learning rate of classifiers. Our proposed method combined several weak classifiers to obtain a suitable boundary for separating data between classes, which could be used to modify classifiers that were incorrectly classified. It should be noted that the Skip-Gram model was also used in our proposed method as a representation technique. The results showed that the use of AdaBoost algorithm has efficiently increased the accuracy of classification. The proposed method was able to obtain the convergence point after 60 epochs with accuracies of 61.25%, 61.93%, 59.02%, 60.16% 64.63 % for extraversion, neuroticism, agreeableness, openness to experience, and conscientiousness respectively.

Employing the combination of photos, videos, and other shared content besides text for predicting individuals' personalities can be considered as possible future works. Moreover, the proposed method of this paper can be utilized in other applications of cognitive science including identifying the level of stress, anxiety, and depression.